\documentclass{article}
\usepackage{ijcai24}
\usepackage{float}
\usepackage{times}
\usepackage{soul}
\usepackage{url}
\usepackage{hyperref}
\hypersetup{hidelinks}
\usepackage[utf8]{inputenc}
\usepackage[small]{caption}
\usepackage{graphicx}
\usepackage{amsmath}
\usepackage{amsthm}
\usepackage{booktabs}
\usepackage{algorithm}
\usepackage{algorithmic}
\usepackage[switch]{lineno}
\usepackage{multirow}
\usepackage{amssymb}

\title{Embedding Ontologies via Incoprorating Extensional and Intensional Knowledge}

\author{
Keyu Wang$^1$\and
Guilin Qi$^1$\and
Jiaoyan Chen$^{2}$\and
Yi Huang$^3$\And
Tianxing Wu$^1$\\
\affiliations
$^1$School of Computer Science and Engineering, Southeast University, Nanjing 211189, China\\
$^2$Department of Computer Science, The University of Manchester, Manchester M13 9PL, UK\\
$^3$China Mobile Research Institute, Beijing 100053, China\\
}
\begin{document}

\maketitle

\begin{abstract}
Ontologies contain rich knowledge within domain, which can be divided into two categories, namely extensional knowledge and intensional knowledge. Extensional knowledge provides information about the concrete instances that belong to specific concepts in the ontology, while intensional knowledge details inherent properties, characteristics, and semantic associations among concepts. However, existing ontology embedding approaches fail to take both extensional knowledge and intensional knowledge into fine consideration simultaneously. In this paper, we propose a novel ontology embedding approach named EIKE (\textbf{E}xtensional and \textbf{I}ntensional \textbf{K}nowledge \textbf{E}mbedding) by representing ontologies in two spaces, called extensional space and intensional space. EIKE presents a unified framework for embedding instances, concepts and their relations in an ontology, applying a geometry-based method to model extensional knowledge and a pretrained language model to model intensional knowledge, which can capture both structure information and textual information. Experimental results show that EIKE significantly outperforms state-of-the-art methods in three datasets for both triple classification and link prediction, indicating that EIKE provides a more comprehensive and representative perspective of the domain.
\end{abstract}

\section{Introduction}

Ontologies, such as YAGO \cite{8}, WordNet \cite{9} and BioPortal \cite{10}, serve as pivotal tools for organizing, representing, and semantically linking heterogeneous information. They provide a structured framework that defines concepts and their relationships, offering a structured knowledge representation that supports various applications like information retrieval \cite{11} \cite{12} and semantic web technologies \cite{31}\cite{32}\cite{llm-dl}. 
\begin{figure}[htbp]
    \centering
   \includegraphics[width=1.1\columnwidth,height=0.55\linewidth]{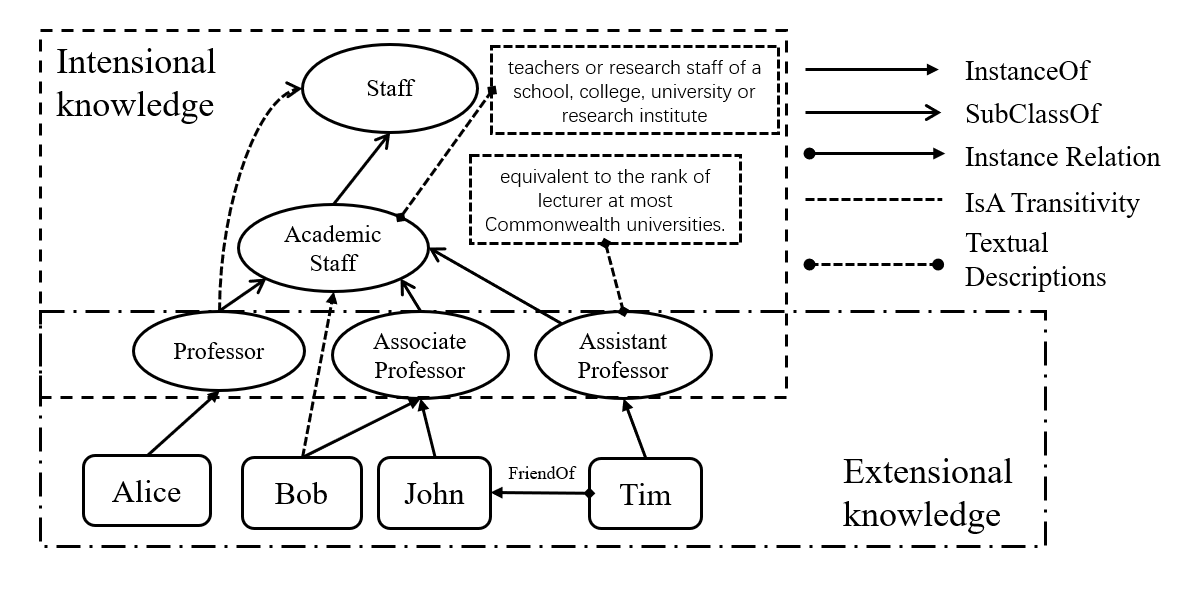}
    \caption{An ontology with extensional and intensional knowledge}
    \label{fig1}
\end{figure} 
Within an ontology, knowledge can be divided into two categories, intensional knowledge, acting as a terminology or taxonomy, describing general properties, characteristics of concepts and their semantic relationships, and extensional knowledge, also known as assertional knowledge, which pertains to specific individuals within the domain and provides type information about the concrete instances that belong to a specific concept \cite{2} \cite{32}. Figure \ref{fig1} shows a sketch of such an ontology. Comprehensively representing both extensional knowledge and intensional knowledge within an ontology is essential for a nuanced, accurate and adaptable understanding of the domain \cite{32}.

Ontology embedding involves transforming each component in an ontology into continuous and low-dimensional vectors while preserving their semantic relationships. Recently, ontology embedding techniques have been widely investigated and applied to various applications such as semantic search and reasoning \cite{13}.  Integrating both extensional knowledge and intensional knowledge enriches ontology embedding and provides a more comprehensive and representative perspective of the domain, resulting in embeddings that better capture semantic relationships and similarities among instances and concepts, as well as better apply to applications such as materialization and classification. 
However, existing ontology embedding models fail to take both of these two types of knowledge into fine consideration simultaneously. For example,  existing geometry-based approaches such as \cite{3}\cite{4} represent extensional knowledge well, but perform poorly in capturing intensional knowledge since they ignore the inherent characteristics and properties of the concepts. Especially, they can't capture lexical and textual information. Concept2Vec \cite{27} utilizes a random walk strategy to learn concept embedding through the similarity between concepts, ignoring the influence of a large number of instances. OWL2Vec* \cite{14} considers the semantic aspects of graph structures, lexical information, and logical constructors, making it suitable for modeling intensional knowledge within an ontology. However, despite the fact that OWL2Vec* takes instances into account, its random walk strategy fail to clearly distinguish the differences between concepts and instances, making it hard to sufficiently model extensional knowledge. 

%\cite{32} defined a concept as a pair $(I, E)$, where $I$ is the intension of the concept and $E$ is the extension of the concept, and found that combining extension and intension of the concepts benefits taxonomy building. This motivates us to consider these two representations of concepts (extension representation and intension representation) and 

In this paper, we propose an innovative method EIKE that delineates ontology representation learning in two distinct spaces, referred to as extensional space and intensional space. The extensional space primarily models the extensional knowledge, geometrically representing concepts as spatial regions and instances as point vectors. The intensional space models the intensional knowledge, applying a pre-trained language model to encode intensional knowledge from properties and characteristics of the concepts. The combination of structure-based embeddings obtained by geometric method in extensional space and text-based embeddings obtained by pre-trained language model in intensional space can fully represent ontologies. After acquiring representations of concepts and instances in extensional space and intensional space, distinct loss functions are proposed for \textit{SubClassOf}, \textit{InstanceOf}, and \textit{Relational} Triples, culminating in joint training.  Experimental results on YAGO39K, M-YAGO39K and DB99K-242 show that EIKE significantly outperforms state-of-the-art methods of triple classification and link prediction tasks. 

Our contributions can be summarized as follows: \\
$ \bullet $ We are the first to consider embedding  extentional knowledge and intensional knowledge into different spaces and propose a framework for joint representation learning of extensional knowledge and intensional knowledge. \\
$ \bullet $	We propose a novel method to utlize a pre-trained language model to encode the intensional knowledge and use the embeddings of intensional knowledge for computing semantic relationships among concepts. \\
$ \bullet $	We conduct experiments on YAGO39K, M-YAGO39K and DB99K-242 to evaluate the effectiveness of EIKE. Experimental results in link prediction and triple classification tasks indicate that our approach outperforms the state-of-the-art methods.

\section{Related Work}

\subsection{Instance-level ontology embedding}
Early knowledge embedding methods,  called factual knowledge graph embedding, treat both concepts and instances as instances. Mainstream models are mainly divided into translational distance models such as TransE \cite{15}, TransH \cite{16}, TransR \cite{17}, TransD \cite{18}, and bilinear models, such as RESCAL \cite{19}, HolE \cite{20}, DistMult \cite{21} and ComplEx \cite{22}.  

Some recent works apply pre-trained language models (PLMs) to knowledge embedding to capture textual information. For example, PretrainKGE \cite{pretrainkge} utilizes BERT \cite{bert} as encoders to derive embeddings of the entities and relations with descriptions, and input them into structured-based knowledge graph embedding models for training.  KEPLER \cite{25} uses PLMs as encoders to derive description-based embeddings and is trained on the objectives of knowledge embedding (KE) and masked language modeling (MLM).   \cite{kgc-encoder} \cite{kgc-s2s} both apply sequence-to-sequence framework and pretrained language models to Knowledge Graph Completion (KGC). More recently, \cite{llm-kgc} explores Large Language Models for KGC. 
These methods can leverage descriptive information in  knowledge graphs but cannot sufficiently represent concepts and model  hierarchical structure.

\subsection{Concept-level ontology embedding}
Concept-level ontology embedding methods consider both instances and concepts in an ontological knowledge base as concepts. For instance, EL Embedding \cite{7} constructs specific scoring functions and loss functions from logical axioms from EL++ by transforming logical relations into geometric relations. Onto2Vec \cite{6} and OPA2Vec \cite{26} are two ontology embedding algorithms based on word embeddings using skip-gram or CBOW architectures. Concept2Vec \cite{27} utilizes a random walk strategy to learn concept embeddings through the similarity between concepts. CosE \cite{5} utilizes angle-based and distance-based approaches to design two scoring functions to learn \textit{SubClassOf} and \textit{DisjointWith} while preserving transitivity and symmetry. Concept-level ontology embedding focuses on ontological concepts, disregarding instances within the knowledge base, hence its limited scope. Although some ontology embedding methods like OWL2Vec* \cite{14} take instances into account, they fail to distinguish concepts and instances clearly because of the nature of their learning strategies like random walk. In contrast, our primary focus revolves around methodologies that differentiate  concepts and instances.

\subsection{Differentiating concepts and instances for ontology embedding}
Recently, some studies have attempted to differentiate between concepts and instances in knowledge representation learning. TransC \cite{3} represents each concept as a spherical region and each instance as a point vector in the same vector space, capable of handling the isA transitivity. TransFG \cite{28} adopts TransC's geometric modeling approach but embeds instances and concepts into different vector spaces to address instance ambiguity. To address the anisotropy issue in TransC and TransFG, \cite{4} introduces TransEllipsoid and TransCuboid, representing concepts respectively as ellipsoidal and cuboidal regions. Additionally, JOIE \cite{29} and DGS \cite{dgs} learn embeddings from two-view knowledge graphs, yet they lacks interpretability for the isA transitivity. \cite{iri} and RMPI \cite{rmpi} show utilizing ontology-view information can enhance instance-view knowledge representation for applications like knowledge graph completion and inductive relation inference.

Different from previous methods like TransFG \cite{28} and JOIE \cite{29} which embed ontologies in two spaces, one space only for concepts and the other only for instances, EIKE embeds extensional knowledge and intensional knowledge in two distinct spaces respectively. This approach represents ontologies more sufficiently since it combine structure-based embeddings and textual-based embeddings. Additionally, EIKE is the first to finely consider two aspects of semantics of a concept: (1) a hight level-summary of its instances; (2) its inherent properties, characteristics and associations with other concepts.

\section{Task Definition}
In this section, we formulate the task. At first, we introduce some preliminaries about the ontology.

\textbf{Ontology.} Based on the problem formulation in \cite{3} \cite{4}, we incorporate considerations for literals and descriptions. We consider the relation between concepts (i.e., \textit{SubClassOf}), the relation between concepts and intances (i.e., \textit{InstanceOf}), and the relations between instances (for convinience, we call them instance relations), and textual descriptions in ontologies.  Ontology is simply formalized as $ O = \{ C,I,R,T,L \} $. $C$ and $I$ denote the sets of concepts and instances respectively. Relation set $R$ is categorized into three types as $R = \{r_e, r_c\} \cup R_l$, where $r_e$ and $r_c$ denote \textit{InstanceOf} relation and \textit{SubClassOf} relation respectively, and $R_l$ is the relation set containing instance relations. The triple set $T$ is divided into three disjoint subsets: \textit{Relational} triple set $T_l = \{(i_m,r_l,i_n) \mid   i_m,i_n \in I,r_l \in R_l\} $, \textit{SubClassOf} triple set $T_c  = \{(c_m,r_c,c_n) \mid c_m,c_n\in C \}$ and \textit{InstanceOf} triple set $T_e  =\{(i,r_e,c) \mid i \in I, c\in C \}$. $L$ denotes the set of  existing descriptions about certain concepts in ontology. 

Typically, an ontology contains two components, intensional knowledge (also called terminological knowledge) and extensional knowledge (also called assertional knowledge), where terminological knowledge contains concepts (eg. \textit{Person}, \textit{Pet}, \textit{Animal}) and their relationships (eg. \textit{Pet SubClassOf Animal}) and assertional knowledge contains instances (eg. \textit{John}, \textit{Fido}), concepts, class assertion (eg. \textit{John InstanceOf Person}, \textit{Fido InstanceOf Pet}) and property assertions (eg. \textit{John hasPet Fido}).

%In this study, we embed ontologies considering extension and intension of the concepts. Despite the fact that the notions of extension and intension are widely used in various areas such as biomedical science \cite{33}, philosophy \cite{36} and linguistics \cite{34}, we follow \cite{1} \cite{2} \cite{32}  for the definition and interpretations. 

%\textbf{Extension.}  The extension of a concept is defined as the set of individuals or instances that belong to this concept. For example, the extension of the concept \textit{President of the United States} is a collection of previous American presidents such as \textit{Joe Biden}, \textit{Donald Trump} and \textit{Barack Obama}.

%\textbf{Intension.}  The intension of a concept refers to its definition or the set of characteristics, properties, or criteria that define or describe this concept. For example, the intension of the concept \textit{President of the United States} includes \textit{"head of state and head of government of the United States of America, who directs the executive branch of the federal government and is the commander-in-chief of the United States armed forces$ \cdots\cdots$"} and other characteristics and properties.

%Basically, the assertional knowledge contains extensional knowledge—knowledge that is specific to the individuals of the domain of discourse. The schematic knowledge contains intensional knowledge in the form of a terminology and is built through declarations that describe general properties of concepts \cite{2}. 

\textbf{Task Definition.} We embed ontologies in extensional space and intensional space. Given an ontology $O$, we map each instance $i$ into a point vector $\mathbf{i}^e \in \mathbb{R}^d$ and each concept c into a geometric region ${G}(\mathbf{c},\mathbf{b})$ in extensional space and map $\mathbf{i}^e$ into a point vector $\mathbf{i}^i \in \mathbb{R}^d$ and concept $c$ into a point vector $\mathbf{c}^i \in \mathbb{R}^d$ in intensional space.  Additionally, the designed approach is expected to retain isA transitivity \cite{3} to some degree, namely, \textit{InstanceOf-SubClassOf}: 
$ (i, r_e, c_1) \in T_e \wedge (c_1, r_c, c_2) \in T_c \to (i, r_e, c_2) \in T_e $  and \textit{SubClassOf-SubClassOf}: $ (c_1, r_c, c_2) \in T_c \wedge (c_2, r_c, c_3) \in T_c \to (c_1, r_c, c_3) \in T_c $. 

\begin{figure*}[htbp]
    \centering
   \includegraphics[width=2.0\columnwidth,height=0.39\linewidth]{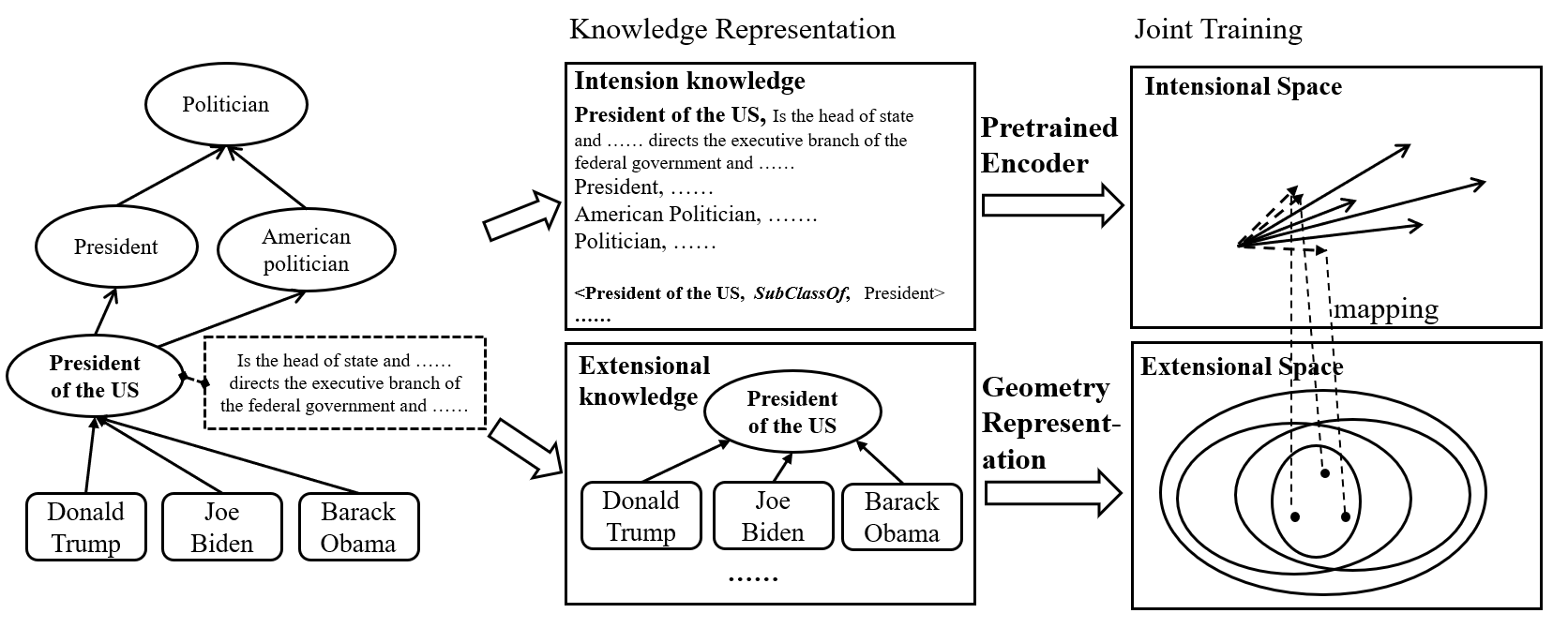}
    \caption{The framework for representing an ontology in extensional space and intensional space}
    \label{fig2}
\end{figure*} 

\begin{figure*}[h]
    \centering
   \includegraphics[width=2.0 \columnwidth,height=0.24\linewidth]{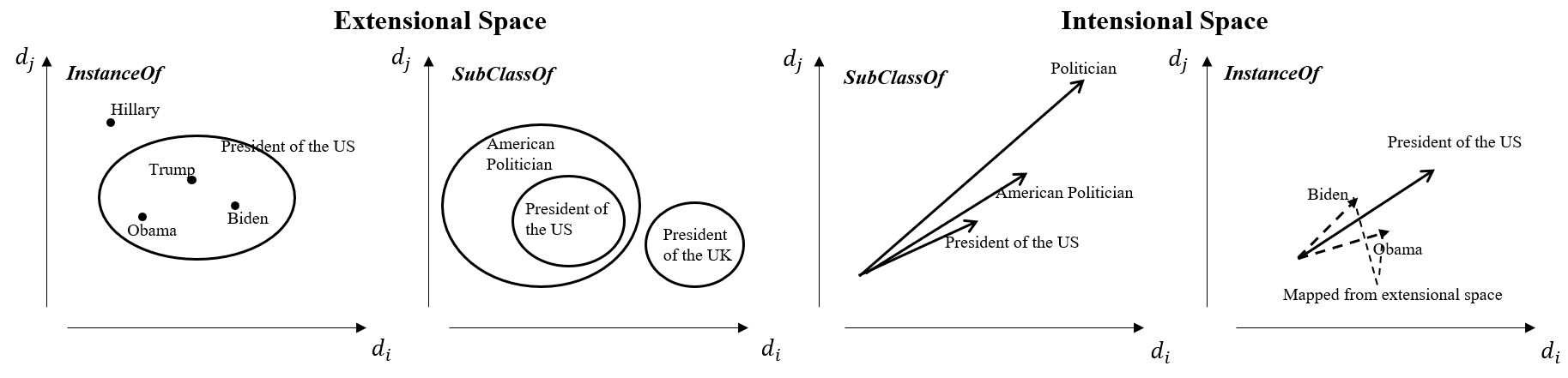}
    \caption{Embedding structure in extensional space and intensional space.}
    \label{fig3}
\end{figure*} 

\section{Method}
In this part, we first introduce the framework for representing instances, concepts and their relations in an ontology. As shown in Figure \ref{fig2}, we introduce two spaces to represent the extensional knowledge and the intensional knowledge within an ontology respectively. In these two spaces, we define scoring functions for modeling three kinds of relations, namely \textit{SubClassOf} triples, \textit{InstanceOf} triples, and \textit{Relational} triples. Finally, we give a method for jointly training ontological concepts and instances.

\subsection{Knowledge representation framework}
 We represent ontology in two spaces, called extensional space and intensional space respectively. Actually, ontological knowledge base contains extensional knowledge and intensional knowledge, and designing two spaces to model these two kinds of knowledge respectively is beneficial to fully capture the information and semantics of the ontology.

\subsubsection{Ontology representation in extensional space}
Extensional knowledge provides the information about the types of instances and relationships between instances. A natural idea of modeling extensional knowledge is to use geometric regions to represent concepts, and point vectors to represent instances. For a concept $c$ and its instances ${i_1, i_2, \cdots, i_n}$, the instance embeddings can be modeled by the vectors $\mathbf{i_1}, \mathbf{i_2}, \cdots, \mathbf{i_n}$ being inside $c$'s region embedding $\mathbf{G}_c$ \cite{3}. This kind of geometry-based method also performs well in capturing \textit{SubClassOf} relation which belongs to intensional knowledge, but it can’t capture more nuanced and comprehensive information of intensional knowledge. In this work, we follow \cite{4} to represent concept as ellipsoid and instance as point vector in the extensional space, shown in the left part of Figure \ref{fig3}. This representation method considers the anisotropy of concept embedding and has a good capability to represent ontological concepts \cite{4}. For each concept $c \in C$, we represent it as ellipsoid $G_c(\mathbf{c},\mathbf{b})$ in extensional space: 
\begin{center}
$ G_c(\mathbf{c}, \mathbf{b}) = \{(x_1, x_2, \ldots, x_d) \mid \sum_{j=1}^d (\frac{x_j - c_j}{b_j})^2 \le 1\}
$ 
\end{center}
where $ \mathbf{c} \in \mathbb{R}^d $ denotes the center point and each dimension of $ \mathbf{b} \in \mathbb{R}^d $ is the each semi-axis length along the corresponding dimension. 

We use fixed-length vectors to represent the instances and the relations between instances, and directly use embedding matrices for initialization, namely:
\begin{gather*}
\mathbf{i}^e  =\mathbf{E}_{\boldsymbol{i}} \\
\mathbf{r}^e  =\mathbf{T}_{\mathbf{r}} 
\end{gather*} 
where $\mathbf{E} \in \mathbb{R}^{|I| \times d}$ is instance embedding matrix, $\mathbf{T} \in \mathbb{R}^{|R_l| \times d}$ is instance relation embedding matrix.

\subsubsection{Ontology representation in intensional space}
Intensional knowledge describes general properties, characteristics of concepts and their semantic relationships. Typically, texts contain abundant intensional information related to the concepts.  Pre-trained language models are trained from large corpus, which can be used to extract rich textual information. This inspires us to use pre-trained language models to encode the intensional knowledge. We input the name\footnote{The term 'name' here generally refers to the name part of the URI of the concepts for practice in real ontologies.} of each concept or concatenation of name and description if description of the concept exists to a pre-trained language model to encode its properties and characteristics. For a concept $c$, we initialize its embedding vector $\mathbf{c}^i$ in intensional space through the pre-trained model: 
\begin{center}
$ \mathbf{c}^i= E(text_c) $
\end{center}
where $\mathbf{c}^i \in \mathbb{R}^d$, $ text_c$ denotes the name or concatenation of name and description of concept $c$, $E$ denotes the encoder of the used pretrained language model. In our experiment,  we replace the underscores in the concept name with spaces and strip off the brackets on both sides to get the text to be input into the encoder. For instance, concept ``\textless wikicat\_Danish\_male\_film\_actors\textgreater"  will be transformed into ”wikicat Danish male film actors”. We use Sentence-BERT \cite{SBERT}, the most popular model for computing text similarity, as the encoder to get the intensional embeddings of ontological concepts. In the experiments, in order to facilitate the comparison, we also conduct experiments of initializing the intensional embeddings of the concepts with random matrix instead of pre-trained model as ablation study. Variants initializing the intensional embeddings directly with random matrix are denoted with UNP, and variants of encoding with pre-trained model are denoted with PRE.

Typically intensional knowledge doesn't contain instances. However, to facilitate joint training, we obtain \textit{virtual} instance embeddings in intensional space by mapping instance embeddings in extensional space: 
$$
\mathbf{i}^i=\mathbf{M}_{e i} \mathbf{i}^{{ }^e}
$$
where $\mathbf{M}_{e i} \in \mathbb{R}^{d \times d}$. In experimental evaluation, we use the parameter-fixed identify matrix and parameter-learnable normal matrix for $\mathbf{M}_{ei}$, which are denoted as EYE and MAT respectively.  We call these embeddings as \textit{virtual} instance embeddings because there aren't independent learnable parameters for each embedding and these embeddings just act as a bridge between extensional space and intensional space.

\subsection{Scoring function}
There are three kinds of triples and we define different loss functions for them respectively.

\textbf{InstanceOf Triple Representation.}  In extensional space, given a triple  $\left(i, r_e, c\right)$, if it is true, the instance embedding vector $\mathbf{i}^e$ should be inside the  region of the corresponding  concept $G_c(\mathbf{c}, \mathbf{b})$. The loss function is defined as:
$$
f_{ins}^{ex}(i, c)=\left[\sum_{j=1}^d\left(\frac{i^e_j-c_j}{b_j}\right)^2-r_c^2\right]_{+}
$$

In intensional space, with the aim to jointly learn knowledge in extansional space and intensional space, for a given triple instance $\left(i, r_e, c\right)$, the loss function in intensional space is defined as:
$$
f_{ins}^{in}(i, c)=1-\cos \left(\boldsymbol{i}^{i}, \boldsymbol{c}^{{i }}\right)
$$

The loss function of \textit{InstanceOf} Triple is defined as:
$$
f_{ins}(i, c)=f_{ins}^{ex}(i, c)+\alpha f_{ins}^{in}(i, c)
$$
where $\alpha$ is a hyper-parameter.

\textbf{SubClassOf Triple Representation.} A given true triple  $\left(c_i, r_c, c_j\right)$ contains two aspects of information: (1) From the perspective of extensional knowledge, the set of $c_i$'s instances is a subset of the set of $c_j$'s instances. In extensional space, $G_{c_i}(\mathbf{c}_i, \mathbf{b}_i)$ is modeled as a sub-area of $G_{c_j}(\mathbf{c}_j, \mathbf{b}_j)$, and the loss function is defined as:
$$
f_{sub}^{ex}\left(c_i, c_j\right)=\left[\sum_{p=1}^d\left(\frac{c_{i p}}{b_{i p}}-\frac{c_{j p}}{b_{j p}}\right)^2+r_{c i}^2-r_{c j}^2\right]_{+}
$$
(2) From the perspective of intensional knowledge,  $c_i$ and $c_j$ have some common characteristics and attributes, and the characteristics and attributes contained in $c_j$ are assumed more abundant than that contained in  $c_i$ since $c_j$ is superordinate concept, more abstract and it may contain other subordinate concepts. In intensional space,  $c_i$ and $c_j$ are embedded with high similarity and the module length of $\mathbf{c_j}^i$  is assumed greater than that of $\mathbf{c_i}^i$. In summary, we define the loss function in intensional space:
$$
f_{sub}^{in}\left(c_i, c_j\right)=1-\cos \left(\mathbf{c}_{\mathbf{i}}^{i}, \mathbf{c}_{\mathbf{j}}^{i}\right)+\left\|\mathbf{c}_{\mathbf{i}}^{i}\right\|-\left\|\mathbf{c}_{\mathbf{j}}^{i}\right\|
$$

In the above formular, $\left\|\mathbf{c}_{\mathbf{i}}^{i}\right\|-\left\|\mathbf{c}_{\mathbf{j}}^{i}\right\|$ can also be used to retain transitivity of concept subsumption in intensional space to some degree \cite{5}.
The loss function of \textit{SubClassOf} Triple can be represented as:
$$
f_{sub}\left(c_i, c_j\right)=f_{sub}^{ex}\left(c_i, c_j\right)+\alpha f_{sub}^{in}\left(c_i, c_j\right)
$$
where $\alpha$ is the same parameter as $\alpha$ in $f_{ins}(i, c)$.

\textbf{Relational Triple Representation.} We define loss function for \textit{Relational} triples only in extensional space. For a relational triple $(h,r,t)$, we utilize the widely used TransE \cite{15} model to calculate the loss function:
$$
f_{rel}(h,r, t)=\|\mathbf{h}^e+\mathbf{r}-\mathbf{t}^e\|_2^2
$$

\subsection{Training method}
For \textit{SubClassOf} triples, we use $\delta$ and $\delta^{\prime}$ to denote a positive triple and a negative triple. $S_{\text {sub }}$ and $S_{\text {sub }}^{\prime}$ are used to describe the positive triple set and negative triple set. Then we can define a margin based ranking loss for \textit{SubClassOf} triples:
$$
L_{sub }=\sum_{\delta \in S_{sub }} \sum_{\delta^{\prime} \in S_{sub }}\left[\gamma_{sub}+f_{sub}(\delta)-f_{sub}\left(\delta^{\prime}\right)\right]_{+}
$$
where $[x]_{+}=\max (0, x)$ and $\gamma_{s u b}$ is the margin separating positive triplets and negative triplets. Similarly, for \textit{InstanceOf} triples and \textit{Relational} triples, The ranking loss will be:
$$
\begin{aligned}
L_{ins} & =\sum_{\delta \in S_e} \sum_{\delta^{\prime} \in S_e^{\prime}}\left[\gamma_{i n s}+f_{ins }(\delta)-f_{ins}\left(\delta^{\prime}\right)\right]_{+} \\
L_{rel} & =\sum_{\delta \in S_{rel}} \sum_{\delta^{\prime} \in S_{r e l}^{\prime}}\left[\gamma_{rel}+f_{rel}(\delta)-f_{rel}\left(\delta^{\prime}\right)\right]_{+}
\end{aligned}
$$

Finally, we define the overall loss function as linear combinations of these three functions:
$$
L=L_{sub}+L_{ins}+L_{rel}
$$

The goal of training our proposed model is to minimize the above function, and iteratively update embeddings of concepts, instances, and relations. Every triple in our training set has a label to indicate whether the triple is positive or negative. We follow \cite{15} to generate negative samples.

\begin{table}[H]
\centering%把表居中
\resizebox{0.5\textwidth}{!}{
\begin{tabular}{cccc}%四个c代表该表一共四列，内容全部居中
\toprule%第一道横线
DataSets & YAGO39K & M-YAGO39K & DB99K-242\\
\midrule%第二道横线 
\text {\#Instance } & 39,374 & 39,374 & 99,744\\
\text {\#Concept } & 46,110 & 46,110 & 242\\
\text {\#Relation } & 39 & 39 & 298\\
\text {\#Training Relational Triple } & 354,997 & 354,997 & 592,654\\
\text {\#Training InstanceOf Triple } & 442,836 & 442,836 & 89,744\\
\text {\#Training SubClassOf Triple } & 30,181 & 30,181 & 111\\
\text {\#Valid (Relational Triple) } & 9,341 & 9,341 &  32,925\\
\text {\#Test (Relational Triple) } & 9,364 & 9,364 &  32,925\\
\text {\#Valid (InstanceOf Triple) } & 5,000 & 8,650 &  4,987\\
\text {\#Test (InstanceOf Triple) } & 5,000 & 8,650 &  4,987\\
\text {\#Valid (SubClassOf Triple) } & 1,000 & 1,187 & 13\\
\text {\#Test (SubClassOf Triple) } & 1,000 & 1,187 & 13\\
\bottomrule%第三道横线
\end{tabular}
}
\caption{{Statistics of YAGO39K, M-YAGO39K and DB99K-242.}}%标题
\label{tab1}
\end{table}

\begin{table*}
\centering%把表居中
\scalebox{0.76}{
\begin{tabular}{c|cccc|cccc|cccc}
\hline Datasets & \multicolumn{4}{|c|}{ YAGO39K } & \multicolumn{4}{c|}{ M-YAGO39K } & \multicolumn{4}{c}{ DB99K-242 } \\
\hline Metric & Accuracy & Precision & Recall & F1-Score & Accuracy & Precision & Recall & F1-Score & Accuracy & Precision & Recall & F1-Score \\
\hline TransE* & 82.6 & 83.6 & 81.0 & 82.3 & $71.0 $ & $81.4 $ & $54.4 $ & $65.2 $ & 78.82  &  79.76 & 77.26 & 78.49 \\
TransH* & 82.9 & 83.7 & 81.7 & 82.7 & $70.1 $ & $80.4 $ & $53.2 $ & $64.0 $ & -- & -- & -- & -- \\
TransR* & 80.6 & 79.4 & ${8 2 . 5}$ & 80.9 & $70.9 $ & $73.0 $ & $66.3 $ & $69.5 $ & -- & -- & -- & --\\
TransD* & 83.2 & 84.4 & 81.5 & 82.9 & $72.5 $ & $73.1 $ & $71.4 $ & $72.2 $ & -- & -- & -- & --\\
HolE* & 82.3 & 86.3 & 76.7 & 81.2 & $74.2 $ & $81.4 $ & $62.7 $ & $70.9 $ & -- & -- & -- & --\\
DistMult* & ${8 3 . 9}$ & ${8 6 . 8}$ & 80.1 & ${8 3 . 3}$ & $70.5 $ & $86.1 $ & $49.0 $ & $62.4 $ & -- & -- & -- & --\\
ComplEx* & 83.3 & 84.8 & 81.1 & 82.9 & $70.2 $ & $84.4 $ & $49.5 $ & $62.4 $ & -- & -- & -- & --\\
Concept2Vec & 81.7 &  81.3 & 82.4 & 81.8 & 69.4 & 78.1 & 56.5 & 59.5 & 67.7 & 63.5 & 79.2 & 55.6 \\
TransC* & 80.2 & 81.6 & 80.0 & 79.7 & 85.3 & 86.1 & 84.2 & 85.2 & 83.59 & 92.38 & 73.21 & 81.96 \\
TransFG* &  80.2 & 82.4 & 78.6 & 80.4 & $85.3 $ & $86.1 $ & $82.2 $ & $85.2 $ & -- & -- & -- & -- \\
JECI* &   83.9 & 86.6 & 83.0 & 84.8 & 86.1 & 88.7 & 84.1 & 86.3 & -- & -- & -- & -- \\
 TransEllipsoid*  & 87.23 & 87.89 & 86.36 & 87.12 & 87.84 & 87.88 & $\mathbf{\underline{87.78}}$ & 87.83 & 93.1 & 94.81  & 91.64 & 93.2 \\
 {  TransCuboid*} & 79.3 & 80.43 & 77.44 & 78.91 & 84.77 & 87.59& 81.03 & 84.18 & 69.91 & 65.23 & 85.28 & 73.92\\
\hline { EIKE-UNP-EYE (unif) } & 87.13 & 86.83 & 87.54 & 87.18 & 87.77 & 89.81 & 85.21 & 87.45 & 92.70 & 94.12 & 91.10 & 92.58\\
 EIKE-UNP-EYE (bern)& 86.81 & 87.26 & 86.20 & 86.20 & 87.57 & 89.12 & 85.60 & 87.32 & $\mathbf{\underline{94.21}}$ & $\mathbf{\underline{96.17}}$ & 92.10 & $\mathbf{\underline{94.09}}$ \\
EIKE-PRE-EYE (unif) & $\mathbf{\underline{89.26}}$ & $\mathbf{\underline{88.81}}$ & $\mathbf{\underline{89.84}}$ & $\mathbf{\underline{89.32}}$ & $\mathbf{\underline{88.80}}$ & 90.08 & 87.20 & $\mathbf{\underline{88.62}}$ & 92.00 & 90.27 & $\mathbf{\underline{94.14}}$ & 92.17 \\
 EIKE-PRE-EYE (bern) & 88.42 & 88.61 & 88.18 & 88.39 & 88.28 & $\mathbf{\underline{9 1 . 3 4}}$ & 84.59 & 87.83 & 92.12 & 92.81 & 91.32 & 92.06\\
EIKE-UNP-MAT (unif) & 85.23 & 86.43 & 84.06 & 85.23 & 85.24 & 86.33 & 83.75 & 85.02 & 78.34 & 72.40 & 91.62 & 80.88\\
EIKE-UNP-MAT (bern) & 81.97 & 83.99 & 79.00 & 81.42 & 83.21 & 89.71 & 75.02 & 81.71 & 78.34 & 72.40 & 91.62 & 80.88\\
EIKE-PRE-MAT (unif) & 84.01 & 85.42 & 82.02 & 83.69 & 80.28 & 85.56 & 72.84 & 78.69 & 85.31 & 82.85 & 89.05 & 85.84\\
 EIKE-PRE-MAT (bern) & 84.34 & 85.76 & 82.36 & 84.02 & 84.64 & 88.58 & 79.54 & 83.82 & 87.75 & 86.10 & 90.03 & 88.02\\
\hline
\end{tabular}}
\caption{Experimental results of \textit{InstanceOf} triple classification(\%). Results of * are taken from \protect \cite{4}}
\label{tab2}
\end{table*}

\section{Experiments}
\subsection{Datasets}
We utilized the same datasets, YAGO39K, M-YAGO39K and DB99K-242 in \cite{4}. YAGO39K was constructed by randomly extracting triples from YAGO, encompassing relations such as \textit{InstanceOf}, \textit{SubclassOf}, and \textit{instance} relations, and M-YAGO39K, on the foundation of YAGO39K, was generated by transitivity utilizing the is-A relation to create additional triples \cite{3}. DB99K-242 \cite{29} is created by removing other relations between concepts except the \textit{SubClassOf} relations  on DB111K-174
 extracted from DBpedia in \cite{4}. Details of YAGO39K, M-YAGO39K and DB99K-242 are presented in Table \ref{tab1}.

\begin{table*}
\centering%把表居中
\scalebox{0.75}{
\begin{tabular}{c|cccc|cccc|cccc}
\hline Datasets & \multicolumn{4}{|c|}{ YAGO39K } & \multicolumn{4}{c}{ M-YAGO39K } & \multicolumn{4}{|c}{ DB99K-242 } \\
\hline Metric & Accuracy & Precision & Recall & F1-Score & Accuracy & Precision & Recall & F1-Score & Accuracy & Precision & Recall & F1-Score\\
\hline TransE* & 77.6  & 72.2 & 89.8 & 80.0 & 76.9  & 72.3 & 87.2 &  79.0 & 61.54 & $\mathbf{\underline{100}}$ & 23.08 & 38\\
TransH* & 80.2 & 76.4 & 87.5 & 81.5 & 79.1 & 72.8 & 92.9 & 81.6 & -- & -- & -- & --\\
TransR* &  80.4 & 74.7 & 91.9 & 82.4 &  80.0 & 73.9 & 92.9 & 82.3 & -- & -- & -- & --\\
TransD* &  75.9 & 70.6 & 88.8 & 78.7 & 76.1 & 70.7 & 89.0 & 78.8 & -- & -- & -- & --\\
HolE* & 70.5 & 73.9 & 63.3 & 68.2 & 66.6 &  72.3 & 53.7 & 61.7 & -- & -- & -- & --\\
DistMult* &  61.9 & 68.7 & 43.7 & 53.4 &  60.7 & 71.7 & 35.5 & 47.7 & -- & -- & -- & -- \\
ComplEx* &  61.6 & 71.5  & 38.6 & 50.1 &  59.8 & 65.6 & 41.4 & 50.7 & -- & -- & -- & --\\
Concept2Vec & 59.3 & 70.3 & 44.5 & 54.8 & 58.9 & 70.4 & 38.5 & 51.9 & 50.4 & 68.4 & 34.57 & 46.2 \\
TransC* & 83.7 & 78.1 & 93.9 & 85.2 &  84.4 & 80.7 & 90.4 & 85.3 & 67.9 & 75 & 23.08 & 35\\
 { TransFG* } & 84.5 & 78.6 & 95.2 & 86.1 & 84.7 & 78.7 & 94.1 & 85.7 & -- & -- & -- & --\\
{ TransEllipsoid*} & 85.1 & 82.1 & 89.7 & 85.7 & 85.5 & 84.2 & 87.4 & 85.8 & 61.54  & $\mathbf{\underline{100}}$ & 23.08 & 38 \\
{  TransCuboid* } & 74.7 & 68.3 & 92.3 & 78.5 & 75.5 & 69.1& 92.4 & 79.1 & 76.92 &  73.33 & 84.62 & $\mathbf{\underline{79}}$ \\
\hline  { EIKE-UNP-EYE (unif) } & 76.80 & 68.77 & $\mathbf{\underline{9 8 . 2 0}}$ & 80.89 & 77.30 & 69.38 & $\mathbf{\underline{9 7 . 7 3}}$ & 81.15 & 53.85 & 57.14 & 30.77 & 40.00\\
{ EIKE-UNP-EYE (bern) } & 76.85 & 69.28 & 96.60 & 80.67 & 77.13 & 69.28 & 97.47 & 80.99 & $\mathbf{\underline{73.08}}$ & 68.75 &84.62 & 75.86\\
 { EIKE-PRE-EYE (unif) } & $\mathbf{\underline{9 0 . 0 0}}$ & $\mathbf{\underline{8 6 . 5 6}}$ & 94.70 & $\mathbf{\underline{9 0 . 4 5}}$ & $\mathbf{\underline{8 7 . 0 2}}$ & $\mathbf{\underline{8 6 . 4 8}}$ & 87.78 & 87.12 & $\mathbf{\underline{73.08}}$ & 65.00 & $\mathbf{\underline{100.00}}$ & 78.79\\
{ EIKE-PRE-EYE (bern) } & 89.05 & 85.60 & 93.90 & 89.56 & 86.81 & 84.33 & 90.47 & $\mathbf{\underline{8 7 . 2 9}}$ & 50.00 & 50.00 & 15.38 & 23.53\\
{ EIKE-UNP-MAT (unif) } & 77.85 & 71.54 & 92.50 & 80.68 & 77.59 & 69.81 & 97.22 & 81.27 & 53.85 & 53.33 & 61.54 & 57.14\\
{ EIKE-UNP-MAT (bern) } & 77.80 & 70.23 & 96.50 & 81.30 & 77.55 & 69.99 & 96.46 & 81.12 & 53.85 & 53.33 & 61.54 & 57.14 \\
{ EIKE-PRE-MAT (unif) } & 77.70 & 71.08 & 93.40 & 80.73 & 77.55 & 70.19 & 95.79 & 81.01 & 53.85 & 54.55 & 46.15 & 50.00\\
{ EIKE-PRE-MAT (bern) } & 77.20 & 70.15 & 94.70 & 80.60 & 76.87 & 70.27 & 93.18 & 80.12 & 61.54 & 58.82 & 76.92 & 66.67\\
\hline
\end{tabular}}
\caption{Experimental results of \textit{SubClassOf} triple classification(\%). Results of * are taken from \protect \cite{4}}
\label{tab3}
\end{table*}

\begin{table*}
\centering%把表居中
\scalebox{0.585}{
\begin{tabular}{c|cc|ccc|cccc|cc|ccc|cccc}
\hline Datasets & \multicolumn{9}{c|}{YAGO39K} & \multicolumn{9}{c}{DB99K242} \\
\hline Experiments & \multicolumn{5}{c|}{ Link Prediction} & \multicolumn{4}{c|}{\textit{Relational}  Triple Classification(\%) } & \multicolumn{5}{c|}{ Link Prediction } & \multicolumn{4}{c}{\textit{Relational}  Triple Classification(\%) }\\
 \hline \multirow{2}{*}{ Metric } & \multicolumn{2}{|c|}{ MRR } & \multicolumn{3}{|c|}{ Hits@N(\%) } & \multirow{2}{*}{ Accuracy } & \multirow{2}{*}{ Precision } & \multirow{2}{*}{ Recall } & \multirow{2}{*}{ F1-Score } & \multicolumn{2}{|c|}{ MRR } & \multicolumn{3}{|c|}{ Hits@N(\%) } & \multirow{2}{*}{ Accuracy } & \multirow{2}{*}{ Precision } & \multirow{2}{*}{ Recall } & \multirow{2}{*}{ F1-Score } \\
 & Raw & Filter & 1 & 3 & 10 & & & & & Raw & Filter & 1 & 3 & 10 & & & &\\
\hline TransE* & 0.114 & 0.248 & 12.3 & 28.7 & 51.1 & 92.1 & 92.8 & 91.2 & 92.0 & 0.17 & 0.232  & 10.2 & 31.5 & 45.7 & 90.5 & 91.5 & 89.29 & 90.38\\
 TransH* & 0.102 & 0.215 & 10.4 & 24.0 & 45.1 & 90.8 & 91.2 & 90.3 & 90.8 & -- & -- & -- & -- & -- & -- & -- & 
 -- & --\\
TransR* & 0.112 & 0.289 & 15.8 & 33.8 & 56.7 & 91.7 & 91.6 & 91.9 & 91.7 & -- & -- & -- & -- & -- & -- & -- & 
 -- & --\\
TransD* & 0.113 & 0.176 & 8.9 & 19.0 & 35.4 & 89.3 & 88.1 & 91.0 & 89.5 & -- & -- & -- & -- & -- & -- & -- & 
 -- & --\\
 HolE* & 0.063 & 0.198 & 11.0 & 23.0 & 38.4 & 92.3 & 92.6 & 91.9 & 92.3 & -- & -- & -- & -- & -- & -- & -- & 
 -- & --\\
 DistMult* & $\mathbf{\underline{0.156}}$ & 0.362 & 22.1 & 43.6 & 66.0 & 93.5 & 93.9 & 93.0 & 93.5 & -- & -- & -- & -- & -- & -- & -- & 
 -- & --\\
 ComplEx* & 0.058 & 0.362 & 29.2 & 40.7 & 48.1 & 92.8 & 92.6 & 93.1 & 92.9 & -- & -- & -- & -- & -- & -- & -- & 
 -- & --\\
 Concept2Vec & 0.098 & 0.344 & 20.4 & 37.2 & 41.7 &  92.4 & 92.6 & 92.8 & 92.6 & 0.132 & 0.158 & 5.8 & 21.3 & 37.2 & 88.6 & 89.8 & 87.4 & 88.9\\
 TransC* & 0.112 & 0.420 & 29.8 & 50.2 & 69.8 & 93.8 & 94.8 & 92.7 & 93.7 & 0.147 & 0.188 & 6.6 & 25.7 & 40.8 &  90.17 & 90.9 & 89.27 & 90.08\\
 { TransFG* }& 0.114 & 0.475 & 32.5 & 52.1 & 70.1 & $\mathbf{\underline{9 4 .4}}$ & 94.7 & 93.7 & 94.0  & -- & -- & -- & -- & -- & -- & -- & 
 -- & -- \\
 { JECI* }& 0.122 & 0.441 & 30.0 & 51.1 & 70.1 & 93.9 &  \textbf{\underline{95.2}} & 93.1 & \textbf{\underline{94.1}}  & -- & -- & -- & -- & -- & -- & -- & 
 -- & -- \\
{ TransEllipsoid*} & 0.112 & 0.536 & 41.6 & $\mathbf{\underline{6 8 . 7}}$ & 75.1 & 92.64 & 92.68 & 92.6 & 92.64 & 0.17 & 0.248 & 11 & 34.4 & 48.1 & $\mathbf{\underline{93.63}}$ & $\mathbf{\underline{94.67}}$ & $\mathbf{\underline{92.46}}$ & $\mathbf{\underline{93.55}}$\\
{TransCuboid*} & 0.095 & 0.475 & 36.3 & 55.6 & 66.9 & 92.18 & 93.8 & 90.26 & 92.02 & 0.149 & 0.200 & 4.2 & 30.4 & 47.8 & 88.4 & 90.26 & 86.1 & 88.13 \\
\hline { EIKE-UNP-EYE (unif) } & 0.093 & 0.563 & 45.6 & 63.8 & 74.5 & 92.69 & 92.54 & 92.87 & 92.70  & 0.167 & 0.247 & 11.1 & 33.8 & 48.3 & 92.38 & 93.46 & 91.14 & 92.29\\
{ EIKE-UNP-EYE (bern) } & 0.115 & $\mathbf{\underline{0 . 5 7 7}}$ & $\mathbf{\underline{4 7 . 1}}$ & 65.4 & $\mathbf{\underline{76.2}}$ & 92.62 & 92.68 & 92.55 & 92.62 & 0.175 & 0.256 & 11.9 & 34.9 & 49.4 & 91.78 & 91.91 & 91.63 & 91.77\\
{ EIKE-PRE-EYE (unif) } & 0.093 & 0.528 & 40.8 & 61.0 & 74.0 & 92.19 & 91.70 & 92.78 & 92.24& 0.164 & 0.243 & 10.9 & 32.9 & 47.7 & 90.42 & 91.61 & 88.99 & 90.28 \\
 { EIKE-PRE-EYE (bern) } & 0.113 & 0.531 & 40.9 & 61.0 & 74.8 & 93.04 & 92.46 & $\mathbf{\underline{9 3 . 7 3}}$ & 93.09 & $\mathbf{\underline{0.181}}$ & $\mathbf{\underline{0.264}}$ & $\mathbf{\underline{12.7}}$ & $\mathbf{\underline{35.3}}$ & $\mathbf{\underline{50.3}}$ & 89.91 & 90.77 & 88.85 & 89.80\\
{ EIKE-UNP-MAT (unif) } & 0.088 & 0.522 & 40.9 & 60.3 & 72.0 & 93.14 & 93.50 & 92.74 & 93.12 & 0.044 & 0.053 & 0.1 & 8.6 & 14.3 & 63.60 & 60.96 & 75.64 & 67.51\\
 { EIKE-UNP-MAT (bern) } & 0.113 & 0.545 & 42.4 & 62.0 & 74.7 & 92.75 & 93.35 & 92.07 & 92.70 & 0.045 & 0.056 & 0.1 & 9.2 & 15.0 & 63.07 & 60.47 & 75.48 & 67.14\\
{ EIKE-PRE-MAT (unif) } & 0.090 & 0.542 & 43.4 & 62.0 & 72.9 & 93.19 & 93.03 & 93.38 & 93.20 & 0.162 & 0.239 & 10.4 & 33.1 & 47.0 & 90.99 & 91.56 & 90.31 & 90.93\\
{ EIKE-PRE-MAT (bern) } & 0.108 & 0.558 & 45.1 & 63.5 & 73.9 & 93.45 & 93.86 & 92.99 & 93.42& 0.172 & 0.252 & 11.4 & 34.9 & 48.4 & 90.47 & 90.82 & 90.03 & 90.43\\
\hline
\end{tabular}}
\caption{Experimental results on link prediction and triple classification for relational triples on YAGO39K and DB99K242. Hits@N uses results
 of “Filter” evaluation setting. Results of * are taken from \protect \cite{4}}
 \label{tab4}
\end{table*}

\subsection{Experimental Settings}
Followed by \cite{4}, we selected 13 state-of-the-art knowledge representation learning models as baselines for evaluating our proposed approach. These models are primarily categorized into two classes: (1) Methods distinguishing concepts and instances, including TransC \cite{3}, TransFG \cite{28}, JECI \cite{jeci}, TransEllipsoid \cite{4} and TransCuboid \cite{4}, (2) methods that do not distinguish between concepts and instances, such as TransE \cite{15}, TransH \cite{16}, TransR \cite{17}, TransD \cite{18}, DistMult \cite{21}, HolE \cite{20}, ComplEx \cite{22} and Concept2Vec \cite{27}. 

Our baselines don't include ontology embedding methods like JOIE and OWL2Vec*. JOIE needs meta-relations of abstract concepts and OWL2Vec* directly embeds ontologies in OWL format and needs both the names of entities and their IRIs. Meta-relations between abstract concepts and IRIs of the entities are not included in the used datasets. 

In this task, we chose the learning rate $\gamma$ for SGD of 0.001, three margins $\gamma_l,\gamma_e,\gamma_c$ from $\{0.1, 0.3, 0.4, 0.5, 1, 2\}$,  $\alpha$ from $\{0.1, 0.5, 1, 2\}$, the number of epoches from $\{1000, 1500, 2000 \}$. The vector dimension was set to $100$. We applied L1 and L2 norms. The optimal configuration for link prediction and triple classification was determined based on Hit@10 and Accuracy metrics on the validation set. The methods 'unif' and 'bern' denote the construction of corrupted triples using uniform distribution and Bernoulli distribution, respectively. We used grid search to determine the hyperparameters for different tasks and datasets. We cover the best configurations that are determined in Appendix. 

All data and codes can be found in supplementary material.

\subsection{Triple Classification}
Triple Classification is a binary task to judge whether a given triple is correct or not. The triple can be an \textit{InstanceOf} triple, a \textit{SubClassOf} triple or a \textit{Relational} triple. The decision rule for classification is simple: We set a relation-specific $\delta_r$ for every relation. For a testing triple, it will be predicted positive if the score is smaller than threshold $\delta_r$, otherwise negative. The relation-specific threshold is optimized by maximizing classification accuracy on the valid set. We train three types of triples \textit{InstanceOf} triples, and \textit{SubClassOf} triples and \textit{Relational} triples simultaneously and show their performances in Table \ref{tab2}, Table \ref{tab3} and Table \ref{tab4} respectively.

From Table \ref{tab2} and Table \ref{tab3}, we conclude that: (1) On all the three datasets, EIKE  performs significantly better than previous methods in \textit{InstanceOf} triple classification and \textit{SubClassOf} triple classification, and EIKE-PRE-EYE (unif) achieves the best performances. This is because EIKE comprehensively considers extensional knowledge and intensional knowledge in ontologies, which can represent concepts more sufficiently.  (2) On YAGO39K and M-YAGO39K, EIKE with pretrained model encoder works better than that without pretrained model encoder in both \textit{InstanceOf} triple classification and \textit{SubClassOf} triple classification, since a pretrained model better encodes the properties and characteristics inherent in concepts and thus enhances the representation learning procedure of intensional knowledge. This is not obvious on DB99K-242, since the intensional knowledge of DB99K-242 is sparse. (3) On YAGO39K and M-YAGO39K, our proposed method using identify mapping matrix perform better than these using normal mapping matrix in both \textit{InstanceOf} and \textit{SubClassOf} triple classification. This may be because parameter-fixed mapping matrix is more suitable for joint learning of extensional space and intensional space. (4) EIKE suffers a slight drop in \textit{InstanceOf} triple classification and a small drop in \textit{SubClassOf} triple classification from YAGO39K to M-YAGO39K. Despite the fact that our methods retain isA transivity in both extensional space and intensional space to some degree, their transitivity effect is not as intuitive as those geometry-based methods only in one space. 

From Table \ref{tab4}, we learn that: For \textit{Relational} triple classification, the performances of our proposed methods are close to that of previous state-of-the-art approach (93.42\% in EIKE-PRE-MAT (bern) and 94.1\% in JECI (sg) in terms of F1-score on YAGO39K, 92.29\% in EIKE-PRE-EYE (unif) and 93.55\% in TransEllipsoid (bern) in terms of F1-score on DB99K-242).

\subsection{Link Prediction}
Link prediction aims to predict the missing head entity (?, r, t) or tail entity (h, r, ?). Same as TransC \cite{3} and \cite{4}, We only tested link prediction on \textit{Relational} triples because the \textit{InstanceOf} relation and \textit{SubClassOf} relation are not suitable for this task. Following most previous works \cite{3} \cite{15} \cite{23}, we adopt two evaluation metrics as followings: (1) the mean reciprocal rank of all correct instances (MRR); (2) the proportion of correct instances that rank no larger than N (Hits@N). Since a corrupted triple may have already existed in knowledge base, which should be regarded as a correct prediction. The settings “Raw” and “Filter’’ distinguish whether or not to consider the impact of a corrupted triple already existing in the knowledge base.

From Table \ref{tab4}, we can find that EIKE performs significantly better especially in terms of Hit@10, Hit@1 and MRR, which indicates modeling extensional knowledge and intensional knowledge in two spaces benefits instance-level embeddings as well.  The “bern” sampling trick and parameter-fixed identify mapping matrix perform well for EIKE.

%\section{Case Study}

\section{Conclusion and Future Work}
In this paper, we proposed a novel ontology embedding model named EIKE. EIKE embeds concepts, instances and relations in two spaces, namely extensional space and intensional space. We utilized a geometry-based method to model the extensional knowledge and applied a pretrained language model to encode intensional knowledge within an ontology. We conducted empirical experiments on YAGO39K, M-YAGO39K and DB99K242. Experimental results show our approach significantly outperforms previous models, indicating comprehensively representing extensional knowledge and intensional knowledge is beneficial for ontology embedding.

For future work, we will explore the following research directions: (1) We will try to establish a unified ontology representation learning framework which can model more relations like \textit{SubPropertyOf}, \textit{Domain} and \textit{Range}. (2) We plan to explore the method with a stronger expressive capability to represent extensional knowledge rather than using spatial region, such as modeling extensional knowledge  by Graph Neural Networks.

\appendix

%\section*{Ethical Statement}

%There are no ethical issues.

\bibliographystyle{named}
\bibliography{ijcai24}

\end{document}